\pdfoutput=1

\documentclass[11pt]{article}

\usepackage[]{EACL2023}

\usepackage{xcolor}
\usepackage{times}
\usepackage{latexsym}
\usepackage{graphicx}

\usepackage[T1]{fontenc}
\usepackage[T5]{fontenc}

\usepackage[utf8]{inputenc}

\usepackage{microtype}

\usepackage{inconsolata}

\usepackage{multirow}
\usepackage[normalem]{ulem}
\useunder{\uline}{\ul}{}
\usepackage{threeparttable}

\usepackage{algorithm2e}

\title{Enriching Biomedical Knowledge for Low-resource Language \\ Through Large-Scale Translation}

\author{Long Phan\textsuperscript{1}$^*$, Tai Dang\textsuperscript{1,3}$^*$, Hieu Tran\textsuperscript{1}$^*$, Trieu H. Trinh\textsuperscript{1,4}$^*$, \\ {\bf Vy Phan\textsuperscript{3}}, {\bf Lam D. Chau\textsuperscript{2}} and {\bf Minh-Thang Luong\textsuperscript{1}}\\
        \textsuperscript{1}VietAI Research\\\textsuperscript{2}Case Western Reserve University\\ \textsuperscript{3}University of Massachusetts-Amherst\\ \textsuperscript{4}New York University}

\begin{document}
\maketitle

\def\thefootnote{*}\footnotetext{The first four authors contributed equally to this work}\def\thefootnote{\arabic{footnote}}

\begin{abstract}

Biomedical data and benchmarks are highly valuable yet very limited in low-resource languages other than English, such as Vietnamese. In this paper, we use a state-of-the-art translation model in English-Vietnamese to translate and produce both pretrained and supervised data in the biomedical domains. Thanks to such large-scale translation, we introduce ViPubmedT5, a pretrained Encoder-Decoder Transformer model trained on 20 million translated abstracts from the high-quality public PubMed corpus. ViPubMedT5 demonstrates state-of-the-art results on two different biomedical benchmarks in summarization and acronym disambiguation. Further, we release ViMedNLI - a new NLP task in Vietnamese translated from MedNLI using the recently public En-vi translation model and carefully refined by human experts, with evaluations of existing methods against ViPubmedT5.

\end{abstract}
\begin{figure*}[ht!]
    \centering
    \includegraphics[width=\textwidth,keepaspectratio]{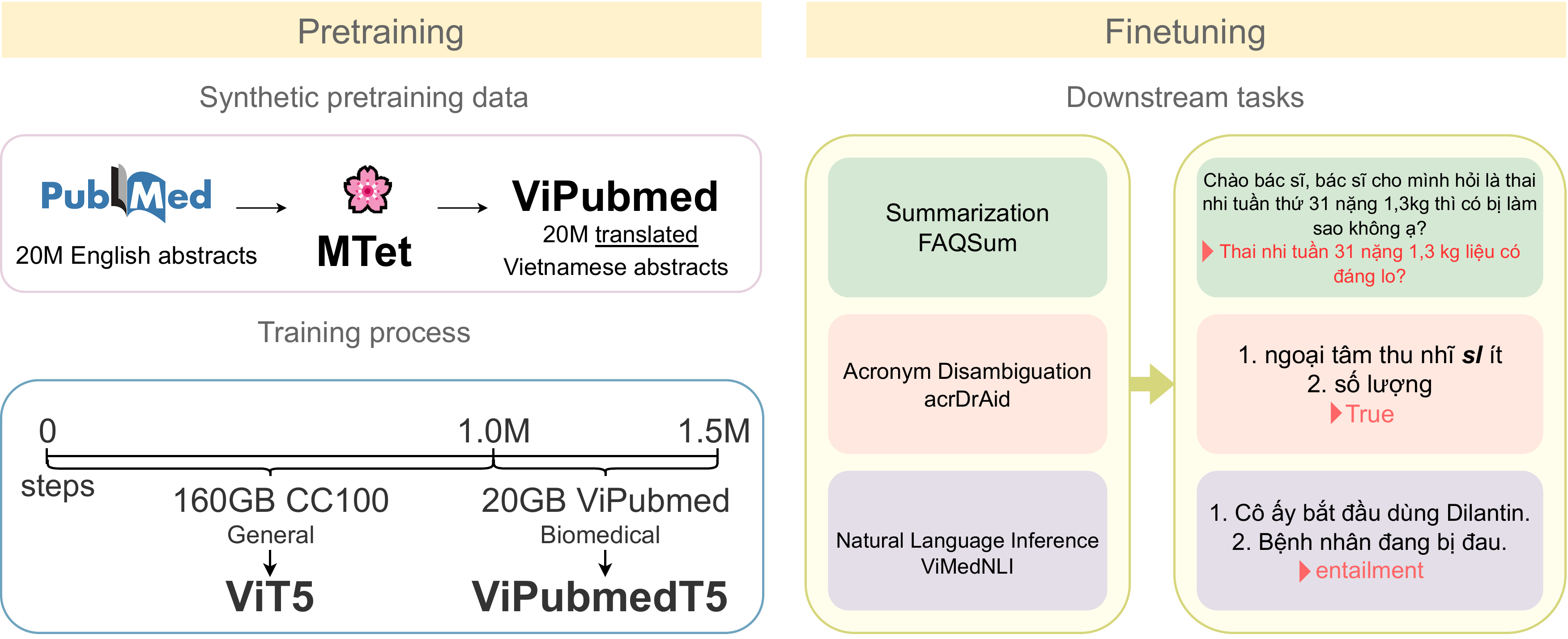}
    \caption{Overview of the pretraining and finetuning of ViPubmedT5}
    \label{fig:vipubmedt5}
\end{figure*}

\section{Introduction}
In recent years, pretrained language models (LMs) have played an important and novel role in developing many Natural Language Processing (NLP) systems. Utilizing large pretrained models like BERT \cite{bert}, XLNET \cite{xlnet}, ALBERT \cite{albert}, RoBERTa \cite{roberta}, GPT-3 \cite{GPT} BART \cite{bart}, and T5 \cite{t5} has become an effective trend in natural language processing. All these large models follow the Transformer architecture proposed by \cite{attention} with the attention mechanism. The architecture has been proven to be very suitable for finetuning downstream tasks leveraging transfer learning with their large pretrained checkpoints. Before the emergence of large Transformer LMs, traditional wording embedding gave each word a fixed global representation. Large pretrained models can derive word vector representation from a trained large corpus. This will give the pretrained model a better knowledge of the generalized representation of a trained language/domain and significantly improve performance on downstream finetune tasks. The success of pretrained models on a generative domain (BERT, RoBERTa, BART, T5, etc.) has created a path in creating more specific-domain language models such as CodeBERT \cite{codebert} and CoTexT \cite{cotext} for coding languages, TaBERT \cite{tabert} for tabular data, BioBERT \cite{biobert} and PubmedBERT \cite{pubmedbert} for biomedical languages.

Biomedical literature is getting more popular and widely accessible to the scientific community through large databases such as Pubmed\footnote{https://pubmed.ncbi.nlm.nih.gov}, PMC\footnote{https://www.ncbi.nlm.nih.gov/pmc}, and MIMIC-IV \cite{mimiciv}. This also leads to many studies, corpora, or projects released to further advance the Biomedical Natural Language Processing field \cite{biobert, pubmedbert, scifive, biobart}. These biomedical domain models leverage transfer learning from pretrained models \cite{bert, electra, t5, bart} to achieve state-of-the-art results on multiple Biomedical NLP tasks like Named Entity Recognition (NER), Relation Extraction (RE), or document classification. 

However, few studies have been on leveraging large pretrained models for biomedical NLP in low-resource languages. The main reason is the lack of large biomedical pretraining data and benchmark datasets. Furthermore, collecting biomedical data in low-resource languages can be very expensive due to scientific limitations and inaccessibility.

We attempt to overcome the issue of lacking biomedical text data in low-resource languages by using state-of-the-art translation works. We start with the Vietnamese language and keep everything reproducible for other low-resource languages in future work. 

We introduce ViPubmedT5, a pretrained encoder-decoder transformer model trained on synthetic Vietnamese biomedical text translated with state-of-the-art English-Vietnamese translation work. Meanwhile, we also introduced ViMedNLI, a medical natural language inference task (NLI), translated from the English MedNLI \cite{mednli} with human refining.


We thoroughly benchmark the performance of our ViPubmedT5 model when pretrained with synthetic translated biomedical data with ViMedNLI and other public Vietnamese Biomedical NLP tasks \cite{vihealthbert}. The results show that our model outperforms both general domain \cite{phobert, vit5} and health-specific domain Vietnamese \cite{vihealthbert} pretrained models on biomedical tasks.  

In this work, we offer the following contributions:
\begin{itemize}
    \item A state-of-the-art English-Vietnamese Translation model (with self-training) on medical and general domains.
    \item A first Encoder-Decoder Transformer model ViPubmedT5 pretrained on large-scale synthetic translated biomedical data.
    \item A Vietnamese medical natural language inference dataset (ViMedNLI) that translated from MedNLI \cite{mednli} and refined with biomedical expertise human.  
    \item We publicize our model checkpoints, datasets, and source code for future studies on other low-resource languages.
\end{itemize}

\section{Related Works}
The development of parallel text corpora for translation and use for training MT systems has been a rapidly growing field of research. In recent years, low-resource languages have gained more attention from the industry, and academia \cite{chen-etal,Shen2021TheSD,gu-etal,nasir}. Previous works include gathering more training data or training large multilingual models \cite{thu-etal, Fan2021BeyondEM}. Low-Language MT enhances billions of people's daily life in numerous fields. Nonetheless, there are specific domains crucial yet limited such as biomedical and healthcare, in which MT systems have not been able to contribute adequately.

Previous works using MT systems for biomedical tasks includes \cite{neves-etal-2016-scielo, neveol-etal-2018-parallel}. Additionally, several biomedical parallel \cite{Delger2009TranslatingMT} have been utilized just for terminology translation only. Pioneer attempts to train MT systems using a corpus of MEDLINE titles \cite{Wu2011StatisticalMT}, and the use of publication titles and abstracts for both ES-EN and FR-EN language pairs \cite{JimenoYepes2012CombiningMA}. However, none of these works targets low-resource languages. A recent effort to train Vietnamese ML systems for biomedical and healthcare is \citet{vihealthbert}. These, however, do not utilize the capability of MT systems, instead relying on manual crawling. Therefore, this motivation has led us to employ MT systems to contribute high-quality Vietnamese datasets that emerged from the English language. To the best of our knowledge, this is the first work utilizing state-of-the-art machine translation to translate both self-supervised and supervised learning biomedical data for pretrained models in a low-resource language setting.



\subsection{Pubmed and English Biomedical NLP Studies}
The Pubmed\footnote{https://pubmed.ncbi.nlm.nih.gov} provides access to the MEDLINE database\footnote{https://www.nlm.nih.gov/bsd/pmresources.html} which contains titles, abstracts, and metadata from medical literature since the 1970s. The dataset consists of more than 34 million biomedical abstracts from the literature collected from sources such as life science publications, medical journals, and published online e-books. This dataset is maintained and updated yearly to include more up-to-date biomedical documents.

Pubmed Abstract has been the main dataset for almost any state-of-the-art biomedical domain-specific pretrained models \cite{biobert, biobart, pubmedbert, biolink, bioelectra, scifive}. In addition, many well-known Biomedical NLP/NLU benchmark datasets are created based on the unlabeled Pubmed corpus \cite{ncbi,pico, ddi, pubmedqa}. 
Recently, to help accelerate research in biomedical NLP, \citet{blurb} releases BLURB (\textbf{B}iomedical \textbf{L}anguage \textbf{U}nderstanding \& \textbf{R}easoning \textbf{B}enchmark), which consists of multiple pretrained biomedical NLP models and benchmark tasks.
It is important to note that all of the top 10 models on the BLURB Leaderboard\footnote{https://microsoft.github.io/BLURB/leaderboard.html} are pretrained on the Pubmed Abstract dataset.

\subsection{English-Vietnamese Translation}
\label{envitranslation}
Due to its limitation of high-quality parallel data available, English-Vietnamese translation is classified as a low-resource translation language \cite{mbart}. One of the first notable parallel datasets and En-Vi neural machine translation is ISWLT'15 \cite{luong-manning-2015-stanford} with 133K sentence pairs. A few years later, PhoMT \cite{phomt} and VLSP2020 \cite{vlsp2020} released larger parallel datasets, extracted from publicly available resources for the English-Vietnamese translation.

Recently, VietAI\footnote{https://vietai.org} curated the largest 4.2M high-quality training pairs from various domains and achieved state-of-the-art on English-Vietnamese translation \cite{mtet}. The work also focuses on En-Vi translation performance across multiple domains, including biomedical. As a result, the project's NMT outperforms existing En-Vi translation models \cite{phomt, m2m100} by more than 2\% in the BLEU score.

\begin{table*}[ht]
\centering
\caption{BLEU Scores Results for En-Vi Translation on MTet Medical and PhoMT General Test Sets}

\begin{threeparttable}

\begin{tabular}{l|l|c|c}
\hline
\textbf{Model}                                             & \textbf{Finetune Datasets}                                                              & \multicolumn{1}{l|}{\textbf{\begin{tabular}[c]{@{}l@{}}MTet Medical \\ Test Set\end{tabular}}} & \multicolumn{1}{l}{\textbf{\begin{tabular}[c]{@{}l@{}}PhoMT General\\ Test Set\end{tabular}}} \\ \hline
M2M100                                                     & CCMatrix + CCAligned                                                                    & 30.18                                                                                              & 35.83                                                                                         \\ \hline
\begin{tabular}[c]{@{}l@{}}Google\\ Translate\end{tabular} & \multicolumn{1}{c|}{-}                                                                  & 38.60                                                                                              & 39.86                                                                                         \\ \hline
SOTA                                                       & MTet+PhoMT                                                                              & 38.69                                                                                          & 45.47                                                                                         \\ \hline \hline
Ours                                                       & \begin{tabular}[c]{@{}l@{}}MTet+PhoMT\\ +1M Self-training Pubmed Abstracts\end{tabular} & \textbf{45.61}                                                                                 & \textbf{46.01}                                                                                \\ \hline
\end{tabular}
\begin{tablenotes}
      \small
      \item \textit{Notes:} The best BLEU scores are in bold. The state-of-the-art (SOTA) model and MTet dataset are from \citet{mtet}; PhoMT dataset and Google Translate's result are from \citet{phomt}. M2M100 model is from \citet{m2m100}.

\end{tablenotes}
\end{threeparttable}
\label{pubmed_selfsupervised_results}
\end{table*}


\section{Improvements on Biomedical English-Vietnamese Translation through Self-training}
\label{self_training_section}

To generate a large-scale synthetic translated Vietnamese biomedical corpus, we first look into improving the existing English-Vietnamese translation system in the biomedical translation domain. Previous work from \citet{mtet} has shown that En-Vi biomedical bitexts are very rare, even for large-scale bitext mining. Therefore, we look into self-training to leverage the available monolingual English biomedical data.

Self-training approach has been experimented with in \citet{selftraining} and utilized to improve translation on low-resource MT systems \cite{fbwat19}. The advantage of this method is that the source side of the monolingual corpus can be domain-specific data for translation. However, the shortcoming is that the generated targets can be low-quality and affect the machine translation performance. Therefore, we start with the English-Vietnamese machine translation model from \citet{mtet}, denoted $bT_A$, which achieves state-of-the-art results on both En-Vi biomedical and general translation domains.

We use $bT_A$ to translate and generate a synthetic parallel biomedical dataset with 1M pairs of English-Vietnamese biomedical abstracts from the Pubmed Corpus. The new 1M En-Vi biomedical pairs are then concatenated with the current high-quality En-Vi translation dataset from MTet \cite{mtet} and PhoMT \cite{phomt}, increasing from 6.2M to 7.2M En-Vi sentence pairs total. To verify the effectiveness of our new self-training data, we re-finetune the $bT_A$ model on this 7.2M bitexts corpus. We report the model performance on the medical test set from MTet and the general test set from PhoMT in Table \ref{pubmed_selfsupervised_results} (the translation performances on other domains like News, Religion, and Law are reported in Appendix \ref{app:selftraining} for further reference).

The results show that our model outperforms existing Machine Translation systems in English-Vietnamese translation by applying self-training. We obtain a significant gain of 6.61 BLEU Score (38.69->45.61) on the MTet Medical test set. Our model with self-training also achieves state-of-the-art results on the PhoMT general domain test set by 0.53 BLEU Score (45.47->46.01). This shows that our approach not only improves the English-Vietnamese translation performance in the biomedical context but also generalizes to general translation. We further discuss our self-training model performance on other translation domains in Appendix \ref{app:selftraining}.

\section{ViPubmed}
\label{vipubmed}
After developing a new state-of-the-art machine translation system for English-Vietnamese translation in the biomedical domain in Section \ref{self_training_section}, we apply the system, denoted $bT_B$, on downstream translation to generate the first large-scale synthetic translated biomedical corpus for Vietnamese.

To ensure that our translated ViPubmed dataset contains up-to-date biomedical research (for example, Covid-19 diseases and Covid-19 vaccines), we use the newest Pubmed22\footnote{https://ftp.ncbi.nlm.nih.gov/pubmed/baseline} which contains approximately 34 million English biomedical abstracts published. The raw data is compressed in XML format. We then parse these structured XMLs to obtain the abstract text with Pubmed Parser\footnote{https://github.com/titipata/pubmed\_parser} \cite{pubmedparser}. 

The machine translation model $bT_B$ is an Encoder-Decoder Transformer based model with 512 token-length for input and output. Therefore, we filter out English Pubmed abstracts with more than 512 tokens. For fair size comparison with the unlabeled dataset of other health-related Vietnamese pretrained models (discussed in Section \ref{baseline}), we take a subset of 20M biomedical abstracts (20GB of text) for translation and leave a larger subset for future releases. We then translate the 20M English biomedical abstracts with the $bT_B$ model using 4 TPUv2-8 and 4 TPUv3-8.

\definecolor{red_custom}{RGB}{233, 116, 81}
\begin{table*}[hbt!]
\centering
\caption{Some Examples of Abbreviations and Spelling Refining}

\begin{tabular}{llll}
\hline
\multicolumn{1}{l|}{\textbf{\#}} & \multicolumn{1}{l|}{\textbf{MedNLI}}                                                                        & \multicolumn{1}{l|}{\textbf{Translated by NMT}}                                                                                      & \textbf{Refined by human}                                                                                       \\ \hline
\multicolumn{4}{c}{\textit{\textbf{Abbreviations Refining}}}                                                                                                                                                                                                                                                                                                                                               \\ \hline
1                                & \multicolumn{1}{l|}{\begin{tabular}[c]{@{}l@{}}Electrocardiograms\\ revealed no\\ \colorbox{green}{QRS} changes\end{tabular}} & \multicolumn{1}{l|}{\begin{tabular}[c]{@{}l@{}}Điện tâm đồ cho thấy\\ không có thay đổi về\\ \colorbox{green}{QRS}.\end{tabular}}                      & \begin{tabular}[c]{@{}l@{}}Điện tâm đồ cho thấy\\ không có thay đổi về\\ \colorbox{green}{phức độ QRS}\end{tabular}          \\ \hline
2                                & \multicolumn{1}{l|}{\begin{tabular}[c]{@{}l@{}}Patient has\\ no \colorbox{green}{PMH}\end{tabular}}                           & \multicolumn{1}{l|}{\begin{tabular}[c]{@{}l@{}}Bệnh nhân\\ không có \colorbox{red_custom}{PMH}\end{tabular}}                                                & \begin{tabular}[c]{@{}l@{}}bệnh nhân \\ \colorbox{green}{không có tiền sử bệnh}\end{tabular}                                 \\ \hline
3                                & \multicolumn{1}{l|}{\begin{tabular}[c]{@{}l@{}}Patient is \\ \colorbox{green}{post op}\end{tabular}}                          & \multicolumn{1}{l|}{\begin{tabular}[c]{@{}l@{}}Bệnh nhân đã\\ \colorbox{red_custom}{hồi phục}\\ \textit{(Patient is \colorbox{red_custom}{recovered})}\end{tabular}}                        & \begin{tabular}[c]{@{}l@{}}Bệnh nhân \\ \colorbox{green}{hậu phẫu thuật}\end{tabular}                                        \\ \hline
\multicolumn{4}{c}{\textit{\textbf{Spelling Refining}}}                                                                                                                                                                                                                                                                                                                                                     \\ \hline
4                                & \multicolumn{1}{l|}{\begin{tabular}[c]{@{}l@{}}The infant was\\ born at \colorbox{red_custom}{herm}\end{tabular}}                  & \multicolumn{1}{l|}{\begin{tabular}[c]{@{}l@{}}Đứa bé được sinh ra\\ ở \colorbox{red_custom}{Herm} \\ \textit{(The baby was born} \\ \textit{at \colorbox{red_custom}{Herm})}\end{tabular}}          & \begin{tabular}[c]{@{}l@{}}Đứa bé được sinh\\ \colorbox{green}{đủ tháng}\\ (The baby was born\\ at \colorbox{green}{term})\end{tabular}        \\ \hline
5                                & \multicolumn{1}{l|}{\begin{tabular}[c]{@{}l@{}}The patient had \\ an \colorbox{red_custom}{sotesophytes}\end{tabular}}             & \multicolumn{1}{l|}{\begin{tabular}[c]{@{}l@{}}Bệnh nhân có\\ \colorbox{red_custom}{sinh cảm}\\ \textit{(The patient has \colorbox{red_custom}{flu})}\end{tabular}}                         & \begin{tabular}[c]{@{}l@{}}Bệnh nhân bị \\ \colorbox{green}{viêm xương khớp}\\ \textit{(The patient had} \\ \textit{\colorbox{green}{osteophytes})}\end{tabular} \\ \hline
6                                & \multicolumn{1}{l|}{Patient has \colorbox{red_custom}{delerium}}                                                                   & \multicolumn{1}{l|}{\begin{tabular}[c]{@{}l@{}}Bệnh nhân có \\ \colorbox{red_custom}{hội chứng delerium}\\ \textit{(Patient has} \\ \textit{\colorbox{red_custom}{delerium syndrome})}\end{tabular}} & \begin{tabular}[c]{@{}l@{}}Bệnh nhân bị \\ \colorbox{green}{mê sảng}\\ \textit{(Patient has \colorbox{green}{delirium})}\end{tabular}                   \\ \hline
\end{tabular}

\begin{tablenotes}
      \item \textit{\#1:} Abbreviation in English can be used in both English and Vietnamese.
      \item \textit{\#2:} Abbreviation can only be used in English. In Vietnamese, abbreviation is different.
      \item \textit{\#3:} Abbreviation  in English is wrong when translated to Vietnamese.
      \item \textit{\#4:} The word \textit{"term"} is misspelled as \textit{"herm"}.
      \item \textit{\#5:} The word \textit{"osteophytes"} is misspelled as \textit{"sotesophytes"}.
      \item \textit{\#6:} The word \textit{"delirium"} is misspelled as \textit{"delerium"}.
\end{tablenotes}
\label{ac_ex_table}

\end{table*}

\section{ViMedNLI}
\label{vimednli}

\begin{figure}[ht!]
    \centering
    \includegraphics[width=0.48\textwidth]{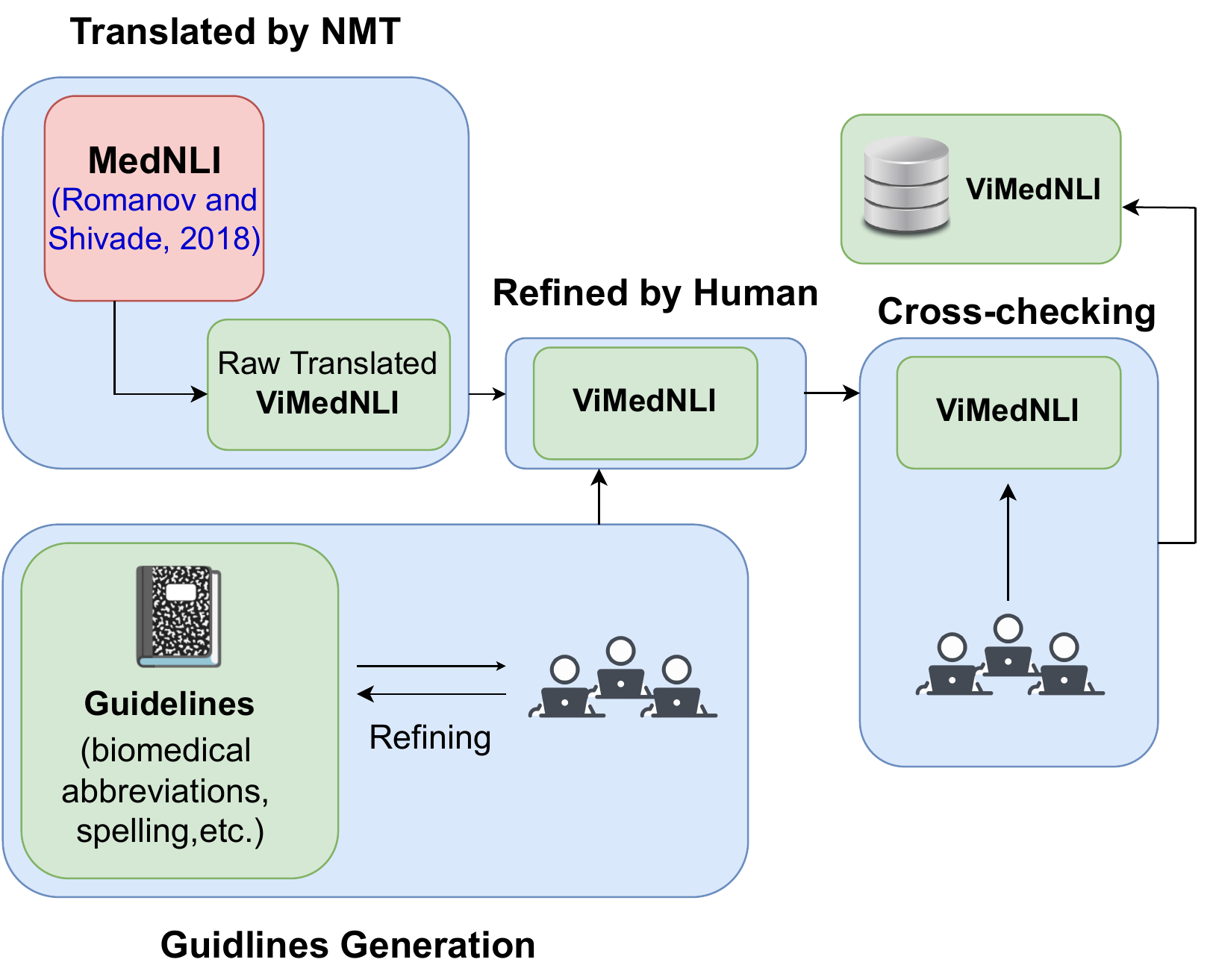}
    \caption{The Process of ViMedNLI Corpus Creation}
    \label{fig:vimednli}
\end{figure}

Along with an unlabeled dataset for pretraining, we also introduce a benchmark dataset generated by translation and refined with human experts. We start with a natural language inference (NLI) task as it is less prone to errors in biomedical entity translation compared to named-entity recognition (NER) or relation extraction (RE) tasks. The process of creating the ViMedNLI is shown in Figure \ref{fig:vimednli}.

\subsection{MedNLI}
MedNLI \cite{mednli} is an NLI dataset annotated by doctors and grounded in the patients' medical history. Given a premise sentence and a hypothesis sentence, the relation of the two sentences (entailment, contradiction, neutral) is labeled by two board-certified radiologists.
The source of premise sentences in MedNLI is from MIMIC-III \cite{mimiciii}, a large open-source clinical database. The dataset has been widely studied and benchmarked by the Biomedical NLP research community\footnote{https://paperswithcode.com/dataset/mednli} \cite{DBLP:journals/corr/abs-1906-05474, scifive, characterbert, bioelectra, kanakarajan-etal-2019-saama}. 


\subsection{Dataset Challenges}
\label{datasetchallenges}
We follow the same procedures discussed in Section \ref{vipubmed} to translate the same training, development, and test sets released in \citet{mednli}. The time and resources to translate the dataset are negligible as there are a total of 14522 samples.

However, upon translating the dataset with NMT, we find out that the English clinical note domain has a distinct sublanguage with unique challenges (abbreviations, inconsistent punctuation, misspellings, etc.). This observation has also been addressed in \citet{FRIEDMAN2002222} and \citet{Meystre2008a}. Such differences in clinical language representation challenge the translation output and our quest to release a high-quality medical dataset.

\begin{table*}[hbt!]
\centering
\caption{Statistics of finetuned datasets}

\begin{tabular}{l|lllll}
\hline
\textbf{Corpus}   & \textbf{Train} & \textbf{Dev} & \textbf{Test} & \textbf{Task}             & \textbf{Domain} \\ \hline
acrDrAid, pairs   & 4000           & 523          & 1130          & Acronym disambiguation    & Medical        \\ \hline
FAQSum, documents & 10621          & 1326         & 1330          & Abstractive summarization & Healthcare            \\ \hline
ViMedNLI, pairs   & 11232          & 1395         & 1422          & Inference                 & Clinical         \\ \hline
\end{tabular}
\label{dataset_stats}

\end{table*}

\subsection{Human Refining}

The unique challenges of clinical data under translation settings (discussed in Section \ref{datasetchallenges}) require us to work with humans who not only have expertise in biomedical knowledge but are also sufficient in both English and Vietnamese languages to refine the dataset. Therefore, we collaborate with pre-medical Vietnamese students who studied at well-known U.S. Universities to refine the ViMedNLI datasets. 

The refining process starts with a comprehensive guidelines document with thorough annotation instructions and examples. Then, as clinical notes contain a significant amount of technical abbreviations that the machine translation system can not translate initially (Section \ref{datasetchallenges}), we work with the medical annotators to create abbreviations and their expansion forms. To make sure the expansion form of these abbreviations generalizes well in real-world settings, we verify the use case of these words through multiple Vietnamese medical websites, blogs, and online dictionaries. Hence, we decided to keep the original English abbreviations, replace them with a Vietnamese expansion form, or replace them with a Vietnamese abbreviation. Some examples of this process are shown in Table \ref{ac_ex_table}. 

Aside from the English medical abbreviations, there are several grammatical and spelling mistakes the machine translation system does not understand, translating either into Vietnamese meanings or even failing to translate. Human refining is therefore required. The phrase \textit{"The infant was born at herm"}, for example, was translated as "Đứa bé được sinh ra ở Herm". The word \textit{"herm"}, which should be spelled as \textit{"term"}, is misspelled and has no medical meaning. The accurate translation should be "Đứa bé được sinh đủ tháng" (\textit{"The infant was born at term"}). Table \ref{ac_ex_table} shows more examples of spelling refining cases. 

Additionally, the machine translation system occasionally produces incorrect Vietnamese meanings when translating words with proper English spelling and grammar. Considering the sentence \textit{"The patient had post-term delivery"} as an example. Despite having the meaning "Bệnh nhân sinh muộn", it was mistranslated as "Bệnh nhân sinh non" (\textit{"The patient had pre-term delivery"}). Another example is "Narrowing of the vessels",  which means "Thu hẹp các mạch" rather than "Thu hẹp các" (no meaning).

\begin{table*}[hbt!]
\centering
\caption{Tests results on Vietnamese health and biomedical tasks}

\begin{tabular}{c|c|cccc|c}
\hline
\multirow{2}{*}{\textbf{Domain}} & \multirow{2}{*}{\textbf{Datasets}} & \multirow{2}{*}{\textbf{Metrics}} & \textbf{PhoBERT} & \textbf{ViT5}  & \textbf{ViHealthBERT}                                            & \textbf{ViPubmedT5}                                               \\ \cline{4-7} 
                                 &                                    &                                   & (+news)          & (+cc100)       & \begin{tabular}[c]{@{}c@{}}(+health \\ text mining)\end{tabular} & \begin{tabular}[c]{@{}c@{}}(+translated \\ ViPubmed)\end{tabular} \\ \hline
Healthcare                          & FAQSum                           & RougeL                           & 41.16             & \textbf{61.3}      &  43.85                                                            & {\ul 60.6}                                                    \\ \hline
Medical                          & acrDrAid                           & Mac-F1                            & 82.51            & {\ul 88}       & 86.7                                                             & \textbf{89.04}                                                    \\ \hline
Clinical                         & ViMedNLI                           & Acc                               & 77.29             & 77.85          & {\ul 79.04}                                                       & \textbf{81.65}                                                    \\ \hline
\end{tabular}
\begin{tablenotes}
      \item \textit{Notes:} The best scores are in bold, and the second best scores are underlined. PhoBERT \& ViHealthBERT scores on FAQSum and acrDrAid are from \citet{vihealthbert}
\end{tablenotes}
\label{results}

\end{table*}

\section{ViPubmedT5}
With an unlabeled synthetic translated ViPubmed Corpus (Section \ref{vipubmed}) and a benchmark ViMedNLI dataset (Section \ref{vimednli}), we pretrain and finetune a Transformer-based language model \cite{attention} to verify the effectiveness of our approach in enriching Vietnamese biomedical domain with translation data. We explain our model and the pretraining settings we applied in this section.

\subsection{Model Architecture}
We adopt the Transformer encoder-decoder model proposed by \citet{attention}, the ViT5 \cite{vit5} checkpoints, and T5 framework \footnote{https://github.com/google-research/text-to-text-transfer-transformer} implemented by \citet{t5}. ViT5 is the first monolingual Vietnamese Transformer model; the model achieves state-of-the-art results on multiple Vietnamese general tasks, including generation and classification. The ViT5 publication releases 2 model sizes - base and large. We train ViPubmedT5 using the base setting (220 million parameters) and leave larger models for future work.

\subsection{Pretraining}
We pretrain our ViPubmedT5 on 20GB of translated biomedical data ViPubmed (Section \ref{vipubmed}). We leverage the Vietnamese checkpoints in the original ViT5 work \cite{vit5} and continuously pretrain the model on the synthetic biomedical-specific data for another 500k steps. Previous works \cite{biobert, pubmedbert} have shown that this approach will allow pretrained language models to learn a better representation of biomedical language context while maintaining the core Vietnamese language representation.

We train ViPubmedT5 using the same spans-masking learning objective as \citet{t5}. During self-supervised training, spans of biomedical text sequences are randomly masked (with sentinel tokens). The target sequence is formed as the concatenation of the same sentinel tokens and the real masked spans/tokens.

\section{Experiments}
\subsection{Benchmark dataset}
We finetune and benchmark our pretrained ViPubmedT5 model on two public Vietnamese biomedical-domain datasets acrDrAid \& FAQSum, \cite{vihealthbert} and our released ViMedNLI (Section \ref{vimednli}). Detailed statistics of the three datasets are shown in Table \ref{dataset_stats}.

\begin{itemize}
    \item \textbf{acrDrAid} \cite{vihealthbert} is a Vietnamese Acronym Disambiguation (AD) dataset that contains radiology reports from Vinmec hospital\footnote{https://vinmec.com/}, Vietnam. The task is correctly identifying the expansion of an acronym in a given radiology report context. The dataset is annotated by three expert radiologists. The acrDrAid has 135 acronyms and 424 expansion texts in total.
    \item \textbf{FAQ Summarization} \cite{vihealthbert} is a Vietnamese summarization dataset collected from FAQ sections of multiple \textit{healthcare} trustworthy sites. For each FAQ section, the question text is the input sequence, and the title is a target summary. 
    \item \textbf{ViMedNLI} is our released dataset discussed in Section \ref{vimednli}.
\end{itemize}

\subsection{Baseline}
\label{baseline}
To verify the effectiveness of our proposed methods, we compare our ViPubmedT5 model with other state-of-the-art Vietnamese pretrained models:

\begin{itemize}
    \item \textbf{PhoBERT} \cite{phobert} is the first public large-scale monolingual language model pretrained for the Vietnamese language. The model follows the original RoBERTa \cite{roberta} architecture. PhoBERT is trained on a 20GB word-level Vietnamese news corpus. 
    \item \textbf{ViT5} \cite{vit5} is the most recent state-of-the-art Vietnamese pretrained model for both generation and classification tasks. The model is trained on a general domain CC100-vi corpus.
    \item \textbf{ViHealthBERT} \cite{vihealthbert} is the first domain-specific pretrained language model for Vietnamese healthcare. After initializing weights from PhoBERT, the model is trained on 25M health sentences mined from different sources.
\end{itemize}

\section{Results}
The main finetuned results are shown in Table \ref{results}. The main takeaway is that training on synthetic translated biomedical data allows ViPubmedT5 to learn a better biomedical context representation. As a result, ViPubmedT5 achieves state-of-the-art in Medical and Clinical contexts while performing slightly worse than ViT5 in healthcare topics.

On the healthcare domain (FAQSum), ViPubmedT5 approximates the current state-of-the-art result (60.6 and 61.3) while outperforming the other models by a large margin (43.85). The slight difference in performance to ViT5 signifies a difference in data distribution in PubMed abstracts (scientific writing) and FAQSum (dialogues between patients and doctors). 

For both medical and clinical datasets, ViPubmedT5 outperforms other existing models. There are also notable improvements from the general domain ViT5 to ViPubmedT5 (88->89.04 in acrDrAid and 77.85->81.65 in ViMedNLI). This indicates that the translated ViPubmed corpus contains biomedical knowledge that low-resource Vietnamese pretrained models can leverage.

Meanwhile, our newly translated ViMedNLI can serve as a robust baseline dataset for Vietnamese BioNLP research. Both health and biomedical domain models (ViHealthBERT \& ViPubmedT5) perform better than general domain models (PhoBERT \& ViT5) on the ViMedNLI dataset. This shows that our translated and refined ViMedNLI dataset is high-quality and has robust biomedical contexts.

\section{Scaling to Other Languages}
Our novel way of utilizing a state-of-the-art NMT system to generate synthetic translated medical data for pretrained models is not limited to the Vietnamese language and is scalable to many other low-resource languages. Recent works focus on improving the quality of multiple low-resource NMT systems \cite{nllb, m2m100, paracrawl}. These new state-of-the-art NMTs make the approach discussed in this paper more practical to produce synthetic translated biomedical data, enriching the Biomedical NLP research knowledge in multiple low-resource languages.

\section{Conclusion}
We utilize the state-of-the-art translation model MTet to scale up the very low-resourced yet highly valuable biomedical data in Vietnamese. Namely, ViPubMedT5, a T5-style Encoder-Decoder Transformer pretrained on a large-scale translated corpus of the biomedical domain that demonstrated state-of-the-art results on both inference and acronym disambiguation in the biomedical domain. We also introduced ViMedNLI, a machine-translated and human-expert refined benchmark in natural language inference to further grow the Vietnamese suite of benchmarks and data in biomedical data.

\section{Limitations}
Although our pretrained model trained on synthetic translated biomedical data produces state-of-the-art results on downstream tasks for the Vietnamese language, the approach hugely depends on the quality of the NMTs for other low-resource languages. Thanks to recent studies and contributions from the Vietnamese research community (Section \ref{envitranslation}), the English-Vietnamese translation system has proven strong enough for us to conduct the experiments discussed in this work. However, the NMT's actual performance needed before making the translated biomedical data useful for pretrained models is still a question that requires further studies.



\section{Acknowledgements}
We would like to thank the Google TPU Research Cloud (TRC) program and Soonson Kwon (Google ML Ecosystem programs Lead) for their supports.

\newpage
\bibliography{anthology,references}
\bibliographystyle{acl_natbib}

\appendix
\newpage
\section{Results for En-Vi Translation with Self-training}
\label{app:selftraining}
We explore a more drastic measure to expand the amount of data in the biomedical domain by self-training. We also explore how a large expansion in the number of biomedical bitexts affects the performance of our model on other domains such as Law, Religion, and News by using the MTet multi-domain test set \cite{mtet}.

\begin{table}[ht]
\centering
\caption{BLEU Scores Results for En-Vi Translation on MTet Multi-domain Test Set}

\begin{threeparttable}

\begin{tabular}{l|clll|}
\hline
\multirow{2}{*}{\textbf{Model}} & \multicolumn{4}{c|}{\textbf{\begin{tabular}[c]{@{}c@{}}MTet Multi-domain\\ Test Set\end{tabular}}}                                \\ \cline{2-5} 
                                & \multicolumn{1}{l|}{Medical}        & \multicolumn{1}{l|}{News}           & \multicolumn{1}{l|}{Religion}       & Laws            \\ \hline
SOTA                            & \multicolumn{1}{c|}{38.69}          & \multicolumn{1}{l|}{\textbf{51.47}} & \multicolumn{1}{l|}{\textbf{41.44}} & 36.43           \\ \hline
Ours                            & \multicolumn{1}{c|}{\textbf{45.61}} & \multicolumn{1}{l|}{51.003}         & \multicolumn{1}{l|}{40.68}          & \textbf{39.51} \\ \hline
\end{tabular}
\begin{tablenotes}
      \small
      \item \textit{Notes:} The best BLEU scores are in bold. The state-of-the-art (SOTA) model and MTet dataset are from \citet{mtet}; Our model trained with self-training approach on an extra 1M En-Vi synthetic biomedical abstracts is discussed in Section \ref{self_training_section}

\end{tablenotes}
\end{threeparttable}
\label{extra_selfsupervised_results}
\end{table}


The improvement is not evident across all domains when tested on a diverse domain test set (MTet). For example, while there are notable improvements in the Medical and Law domain, the model performs worse in the Religion and News domains. This can be attributed to the context representation of biomedical Pubmed Abstract data. Scientific abstracts tend to be more formal and academic for knowledgeable audiences with more domain expertise. Therefore, training on such data allows the Machine Translation system to perform better not only on the trained domain (Medical) but also on other formally presented domains, such as Law, while at the same time performing slightly worse on other domains (News and Religion).   

\end{document}